\documentclass[runningheads]{llncs}
\usepackage[T1]{fontenc}
\usepackage{cite}
\usepackage{graphicx}
\usepackage{enumitem}
\usepackage{multicol}
\usepackage{multirow}
\usepackage{float}
\usepackage{amsfonts,amssymb}
\usepackage{hyperref}
\hypersetup{hypertex=true,
colorlinks=true,
linkcolor=red,
anchorcolor=green,
citecolor=black}
\usepackage[justification=centering]{caption}
\usepackage{makecell}
\usepackage{booktabs}
\usepackage{float}
\usepackage{bbding}
\usepackage{tabu}
\usepackage[labelfont=bf,textfont=normalfont,singlelinecheck=off,justification=raggedright]{caption}
\usepackage{subcaption}
\usepackage{adjustbox}
\usepackage{ifoddpage}
\usepackage{varwidth}
\usepackage{collectbox}

\begin{document}
\title{FIRP: Faster LLM inference via future intermediate representation prediction}

\titlerunning{FIRP: Faster LLM inference}
%
%
\newcommand*\samethanks[1][\value{footnote}]{\footnotemark[#1]}

\author{Pengfei Wu\inst{1}\thanks{\enspace Equal contribution.} \and Jiahao Liu\inst{2}\samethanks \and Zhuocheng Gong\inst{1} \and \\ Qifan Wang\inst{3} \and Jinpeng Li\inst{1} \and Jingang Wang\inst{2} \and Xunliang Cai\inst{2} \and \\ Dongyan Zhao\inst{1,4(}\Envelope\inst{)}}
\authorrunning{P. Wu et al.}
%
\institute{Wangxuan Institute of Computer Technology, Peking University, Beijing, China \and Meituan \and Meta AI \and National Key Laboratory of General Artificial Intelligence \\
\email{\{pengfeiwu1999, lijinpeng\}@stu.pku.edu.cn \\ \{gzhch, zhaody\}@pku.edu.cn \\ \{liujiahao12, wangjingang02, caixunliang\}@meituan.com \\ wqfcr@fb.com}}

\maketitle              
\begin{abstract}
Recent advancements in Large Language Models (LLMs) have shown remarkable performance across a wide range of tasks. Despite this, the auto-regressive nature of LLM decoding, which generates only a single token per forward propagation, fails to fully exploit the parallel computational power of GPUs, leading to considerable latency. To address this, we introduce a novel speculative decoding method named FIRP which generates multiple tokens instead of one at each decoding step. We achieve this by predicting the intermediate hidden states of future tokens (tokens have not been decoded yet) and then using these \textbf{pseudo} hidden states to decode future tokens, specifically, these pseudo hidden states are predicted with simple linear transformation in intermediate layers of LLMs. Once predicted, they participate in the computation of all the following layers, thereby assimilating richer semantic information. As the layers go deeper, the semantic gap between \textbf{pseudo} and \textbf{real} hidden states is narrowed and it becomes feasible to decode future tokens with high accuracy. 
To validate the effectiveness of FIRP, we conduct extensive experiments, showing a speedup ratio of \textbf{1.9x-3x} in several models and datasets, analytical experiments also prove our motivations.

\keywords{Large langurage model  \and Speculative decoding.}
\end{abstract}

\section{Introduction}

Recent developments in Transformer-based large language models~\cite{touvron2023llama,touvron2023llama-2,roziere2023code} have demonstrated remarkable performance across a broad spectrum of tasks. However, these models grapple with excessive inference latency due to the inherently serial process of generating one token per forward propagation. The inference process is typically memory bandwidth-bound, which means most of the inference time is spent loading billions of parameters from memory rather than computing, resulting in the waste of the parallel computational power of GPUs~\cite{shazeer2019fast}. To address this bottleneck, contemporary works propose speculative decoding~\cite{leviathan2023fast, chen2023accelerating, zhang2023draft}, which utilizes a small language model to draft some future tokens, then the LLM verifies and accepts the drafted tokens. While this acceleration technique achieves promising performance, it has its limitations: traditional speculative decoding requires another model to generate draft tokens. 
It is inconvenient to deploy an extra model in some scenarios as it requires more sophisticated scheduling and the draft model might consume extra GPU and memory resources.
Thus, some researchers have been investigating single-model acceleration. That is, to speed up the inference of LLMs without auxiliary models. Self-speculative decoding~\cite{zhang2023draft} and Medusa~\cite{medusa} are typical methods within this line of research.

\begin{figure}[htbp]
\centering  
\begin{subfigure}[b]{0.45\textwidth}
\centering
\includegraphics[width=5cm, height=3cm]{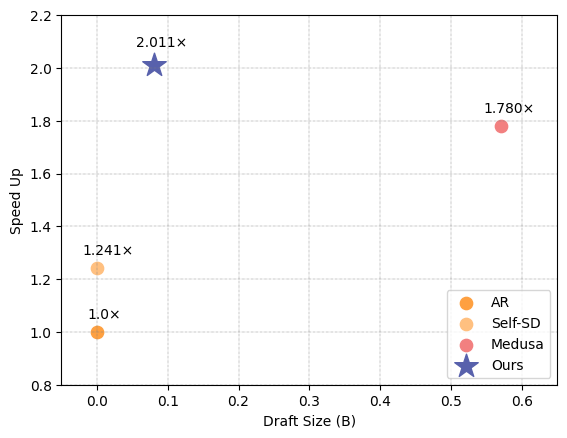}
\caption{End-to-end speedup ratio and draft size}
\label{fig:first_pic1}
\end{subfigure}
\hspace{0.9cm} 
\begin{subfigure}[b]{0.45\textwidth}
\centering
\includegraphics[width=5cm, height=3cm]{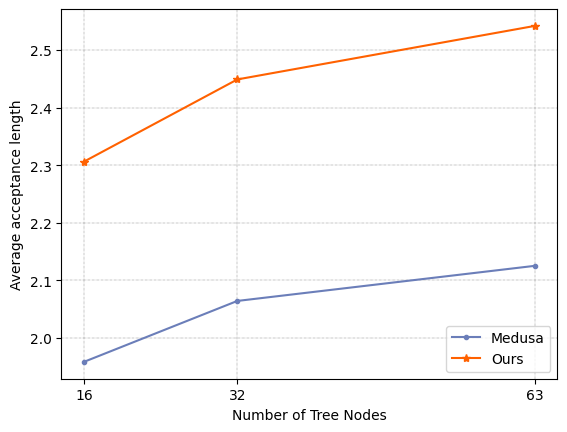}
\caption{Number of average accepted tokens per forward propagation}
\label{fig:first_pic2}
\end{subfigure}
\caption{(a) denotes the end-to-end speedup ratio and draft size for different decoding methods, and (b) denotes the number of average accepted tokens per forward propagation. We search for the best tree structure of Medusa and our FIRP using the search algorithm in~\cite{medusa}. All the results are conducted on LLaMA2-Chat-13B and Xsum dataset and set $k$=3.}
\label{fig:graphical_data}
\vspace{-5mm}
\end{figure}

Our method aligns with the line of single-model acceleration. We design a novel method called \textbf{FIRP}, to predict the hidden states of future tokens in the intermediate layers. We use a simple trainable linear projection to predict the pseudo hidden states of future tokens in a certain intermediate layer, and the pseudo hidden states pass the subsequent layers and interact with the whole sequence to assimilate richer semantic information, in the last layer, we use the original lm-head and decode the draft tokens of the future positions. We outperform baselines in draft size, end-to-end acceleration ratio, and average acceptance length as shown in Figure~\ref{fig:graphical_data}

In this way, we can predict the future draft tokens by predicting their intermediate hidden states in a single forward propagation. 
In addition to the novel design of FIRP, we also use tree attention mechanism~\cite{medusa, miao2023specinfer, spector2023accelerating} to verify multiple draft sequences simultaneously for a better speedup ratio.
The motivation for FIRP is that the pseudo hidden states will interact with themselves and previous hidden states of the context during the forward propagation in which gain more semantic information to boost the accuracy of predicting future tokens.
Our experiments show that this motivation is fulfilled, and our method can achieve better draft token prediction accuracy and inference acceleration ratio compared with other methods under a single-model setting. Our key contributions are:

1. To our best knowledge, we are the first to study the prediction of hidden states of the future tokens in LLMs, our experiments prove that intermediate hidden states could be predicted directly and refined in the forward propagation.

2. We propose FIRP, a novel single-model lossless acceleration method for improving the inference efficiency of LLMs. Our method predicts multiple draft tokens with pseudo hidden states.

3. We conduct various experiments to prove the effectiveness of our method, including some analytic experiments to prove our motivation.


\section{Related Work}
Speculative decoding methods aim to quickly generate some draft tokens and use the LLMs to verify them in parallel to keep lossless generation theoretically. We can divide this class of methods into single model and multiple models approaches. 
The single models' approach is represented by Medusa~\cite{medusa}, Lookahead~\cite{fu2024break}, EESD~\cite{Liu2024SpeculativeDV} and Self-speculative decoding method~\cite{zhang2023draft}. Our method is optimized based on medusa which train extra multiple heads to predict future tokens. Self-speculative methods use a subset of intermediate layers of the whole LLM to generate draft tokens.
The multiple models' approach is represented by traditional speculative decoding~\cite{leviathan2023fast, chen2023accelerating, Gong2024GraphStructuredSD}, which uses an extra small language models~(SLMs) as the draft model to generate draft tokens, the LLMs verify the tokens in parallel.

\section{Methodology}
In this section, we first give some preliminaries, including the overview of traditional auto-regressive decoding algorithm and speculative decoding algorithm, then we provide a detailed description of the training and testing process of FIRP.   

\subsection{Preliminaries}
Language models aim to construct the the ${n+1}_{th}$ token's distribution given the prefix $n$ tokens. Auto-regressive generation method generates one token per model forward propagation. The essence of speculative decoding algorithms, lies in their ambition to increase the expected number of tokens generated in one forward propagation of LLMs while maintain the generation consistency. Thus, the optimization goal of this class of methods can be written to find a maximum positive integer $k$ that satisfies the following conditions:

\begin{equation}
{
\tilde{P}(X_{n+k+1}\ldots X_{n+1}| X_{\leq n}) = P(X_{n+k+1}\ldots X_{n+1}|X_{\leq n})
}
\end{equation}
Where $P$ and $\tilde{P}$ represent the token distribution given by the auto-regressive decoding method and speculative decoding algorithms respectively. For most speculative decoding algorithms, $\tilde{P}$ is obtained by a process including draft stage and verification stage, in the verification stage, tree attention is wildly adopted to verify multiple candidate sequences simultaneously, so a lot of works focus on how to generate better candidates in the draft stage, the draft process can be described as the following formula:

\begin{equation}
{
X_{n+k+1}, \ldots, X_{n+1} = \mathcal M(X_{\leq n})
}
\end{equation}
In some speculative decoding algorithms, $\mathcal M$ represents small language models, or part of the original LLMs~\cite{zhang2023draft}, in the block-wise series of methods (represented by Medusa~\cite{medusa}), $\mathcal M$ can be viewed as the extra heads in the last layer of LLMs, In our method, $\mathcal M$ can be viewed as the linear projection in some intermediate layers, in the following section, we concisely introduce our method, including the training and inference stages.

\begin{figure*}[h]
  \centering
  \includegraphics[width=1\textwidth, height=0.55\textwidth]{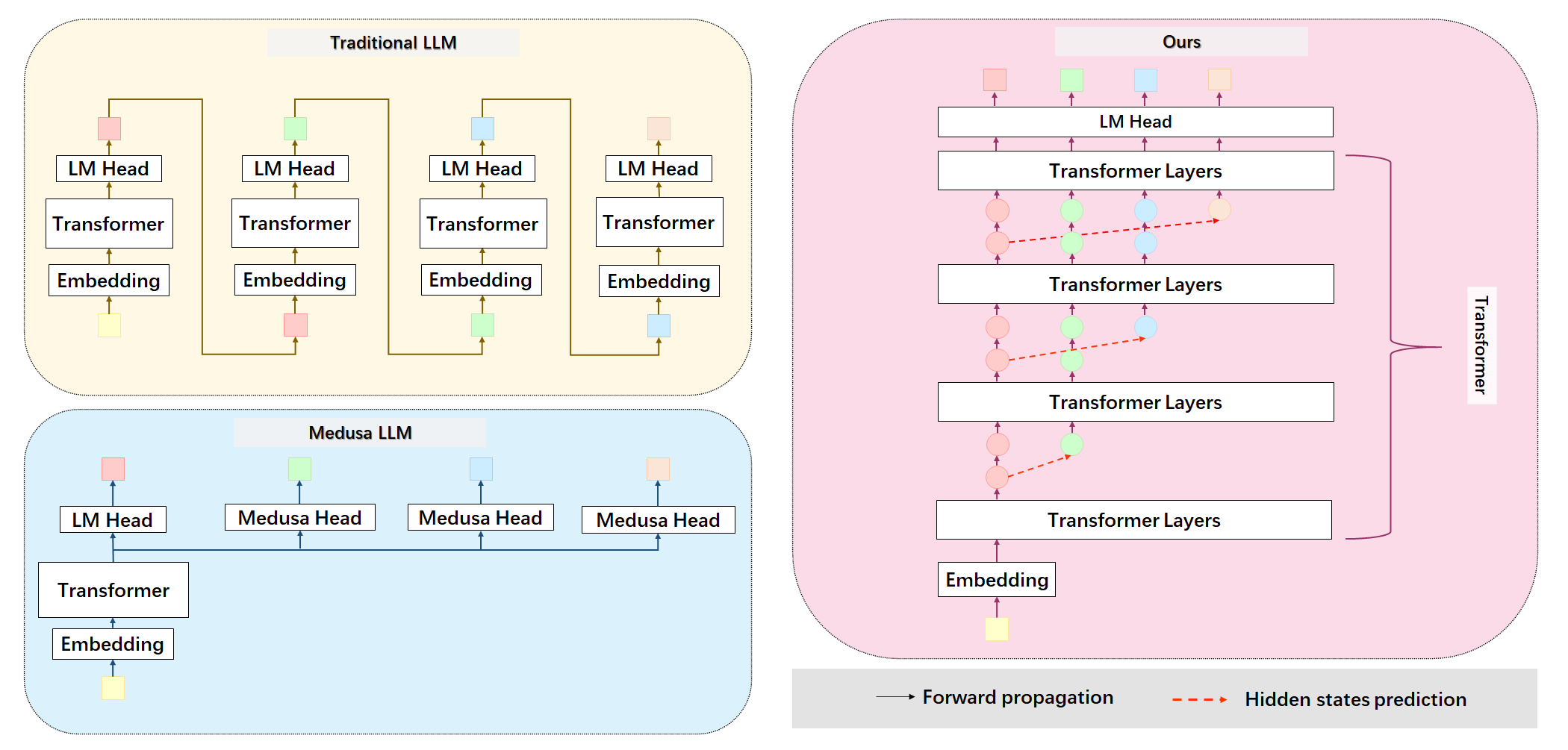}
  \caption{Overview of our method compared with Auto-regressive and Medusa. Our method differs from Medusa because it predicts the intermediate hidden states of future tokens which achieve better prediction accuracy}
  \label{fig:main}
\vspace{-10mm}
\end{figure*}


\subsection{FIRP}
The core idea of our FIRP is to predict the future $k$-gram draft tokens by one step of LLMs' forward propagation, our method choose to train multiple projection (simple linear projections) to \textbf{predict} the future $k$-gram draft token's hidden states in some intermediate layers by mapping the $h_{n}^{j}$ to $h_{n+1}^{j}$ ... $h_{n+k}^{j}$, so we can continue the forward propagation with $n+k$ intermediate hidden states, During the following transformer layers, the last $k$ pseudo hidden states will interact with the whole context and assimilate rich semantic information by attention mechanism, finally they pass the original lm-head to predict the token distribution of future positions normally, the overview of our decoding method compared with Medusa and Auto-regressive is shown in  Figure\ref{fig:main}. 

\subsection{Training stage}
During the training stage, it is required to train multiple linear projections in multiple fixed layers, we denote the number of prediction step as $K$ and the index of prediction layer for step $i$ as $t_{i}$ ($1 \leq i \leq K$). $K$ and $t_{i}$ are considered as hyper-parameters. We train $K$ linear projections separately so we need to conduct the training process $K$ times. For simplicity we discuss the training process of the $i_{th}$ prediction step, we can use $W_{t_{i}}^{i} \in \mathbb{R}^{d \times d}$, $b_{t_{i}}^{i} \in \mathbb{R}^{1 \times d}$ ($1 \leq i \leq K$) to denote the trainable linear projection for the $i_{th}$ step ($d$ represent the hidden dimension of LLMs). Given token sequence $X_{1}$, $X_{2}$, $...$, $X_{n}$, it first conduct the forward propagation to the $t_{i}$ layer to get their corresponding hidden state $h_{1}^{t_{i}}$, $h_{2}^{t_{i}}$, $...$, $h_{n}^{t_{i}}$, then the $n$ hidden states are used to predict their corresponding pseudo hidden states, which can be formulated as below:
\begin{equation}
{
\widetilde{h}_{n}^{t_{i}}, \widetilde{h}_{n-1}^{t_{i}}, \ldots, \widetilde{h}_{1}^{t_{i}} = W_{t_{i}}^{i} \cdot (h_{n}^{t_{i}}, h_{n-1}^{t_{i}}, ...,h_{1}^{t_{i}}) + b_{t_{i}}^{i}
}
\end{equation}
After predicting $\widetilde{h}_{j}^{t_{i}}$ ($1 \leq j \leq n$), we concatenate the original hidden states and pseudo hidden states into a new sequence (i.e.$h_{1}^{t_{i}},..., h_{n-1}^{t_{i}},h_{n}^{t_{i}}, \widetilde{h}_{1}^{t_{i}}, \ldots, \widetilde{h}_{n-1}^{t_{i}}, \widetilde{h}_{n}^{t_{i}}$). It's easy to show that the $\widetilde{h}_{j}^{t_{i}}$ is the pseudo hidden state of $h_{j+i}^{t_{i}}$, so in the self-attention layers it can only view the hidden states from $h_{1}^{t_{i}}$ to $h_{j+i-1}^{t_{i}}$ and itself in the sequence, we design attention mask to achieve the goal. In order to ensure the consistency between the training stage and inference stage, we set the position id $j+i$ to $\widetilde{h}_{j}^{t_{i}}$ to construct the position embedding. 


We then continue to forward the new sequence, and get the final representations in the last layer (i.e., $h_{1}^{l}, \ldots, h_{n-1}^{l}, h_{n}^{l}, \widetilde{h}_{1}^{l}, \ldots, \widetilde{h}_{n-1}^{l}, \widetilde{h}_{n}^{l}$, where $l$ denotes the number of layers of LLMs). We then pass the hidden states sequence through the original lm-head and finally get the token distribution of each position ($P_{1}^{l}, \ldots, P_{n-1}^{l}, P_{n}^{l}, \widetilde{P}_{1}^{l}, \ldots, \widetilde{P}_{n-1}^{l}, \widetilde{P}_{n}^{l}$). We use the KL-divergence as the supervised signal. The loss can be formulated as below:
\begin{equation}
{\small
\mathcal L = \sum_{q=1}^{n-i} \mathcal KL(\widetilde{P}_{q}^{l}, {P}_{q+i}^{l})
}
\end{equation}
\subsection{Inference stage}
In the inference stage, the process is to generate some draft sequences and then verify them to ensure the generation consistency. Multiple draft sequences are constructed into a tree structure by merging their common ancestors, then we flat the tree structure into a tree sequence (denoted as ${T}_{seq}$) and send ${T}_{seq}$ into the LLMs to verify multiple draft sequences simultaneously using tree attention. Each LLM forward propagation we verify ${T}_{seq}$ generated in last round and find the longest draft sequence accepted(denoted as $S$) then generate the new ${T}_{seq}$ on the basis of $S$ at the same time.

\subsubsection{Tree attention}
When predicting the draft tokens for the following several positions, it becomes clear that the draft tokens belonging to different steps form a tree structure according to their order. The tree attention mechanism transforms this hierarchical tree into a linear sequence while preserving the original positional indices of each token. Moreover, it employs a specialized attention mask to ensure that a token only attends to its ancestors within the tree structure, thereby upholding the causal language model's properties. During inference, a pre-defined tree structure and parser facilitate the rapid transformation of token candidates between the tree and sequence representations, obviating the need for additional computations.

\section{Experiment}
\subsubsection{Setup} We evaluate the our method on two different series of models with different size, including LLaMA-2-Chat-13B~\cite{touvron2023llama-2}, LLaMA-2-Chat-7B~\cite{touvron2023llama-2} and Vicuna-13B-v1.5~\cite{vicuna2023}, Vicuna-7B-v1.5~\cite{vicuna2023}. We use greedy sampling strategy and we divide the experiments into main experiments, analytical studies and ablation studies. The main experiments are conducted on all models and the analytical and ablation studies only conducted on 7B model for simplicity.


In the main experiments subsection, we compare our end-to-end time acceleration ratio with Medusa~\cite{medusa}, Lookahead~\cite{fu2024break} and Self-speculative decoding~\cite{zhang2023draft} to show the effectiveness of our method, For Self-speculative decoding method we use the acceleration ratio reported in their paper of LLaMA-2-Chat-13B for the Xsum dataset, and run its open source code to evaluate their effectiveness on the Gsm8K dataset and vicuna-13B model. For Lookahead decoding method, we use the acceleration ratio for MT-bench reported in their paper, For Medusa and our FIRP method, we set $k$=3 and search for the best tree structures as~\cite{medusa} under 16 32 and 63 tree nodes to choose the best results reported, we conduct hidden states prediction on the ${25}_{th}$ ${30}_{th}$ ${35}_{th}$ layers for 13B models and ${15}_{th}$ ${20}_{th}$ ${25}_{th}$ layers respectively.

Analytical experiments and ablation study aim to further prove the effectiveness of our method and verify our motivation. All experiments are conducted on a single NVIDIA A100-80GB GPU and all implementations are based on PyTorch using HuggingFace’s architecture~\cite{wolf2020transformers, lhoest2021datasets}.

\subsubsection{Datasets} We use ShareGPT dataset as our training dataset for all models, and we use the test split of Extreme Sum marization (XSum)~\cite{narayan2018don}, Gsm8k~\cite{cobbe2021training} and MT-bench as our test dataset. ShareGPT is a multi-round conversations dataset comprises nearly 70,000 samples, We train two epoch for all the models. XSum~\cite{narayan2018don} is a dataset for evaluation of abstract single-document summary systems, it's test split has 11,334 samples. we only sample 1000 sentences followed~\cite{zhang2023draft}. The Gsm8k dataset~\cite{cobbe2021training} encompasses a collection of 8,500 linguistically varied, high-quality math word problems for grade school students, all of which were meticulously crafted by human authors, we use its whole test split with 1000 samples. Xum and Gsm8k are evaluated under 1-shot setting followed~\cite{zhang2023draft}.

\subsection{Main Results}
Table~\ref{tab:main experiment xsum gsm8k} and ~\ref{tab:main experiment2} shows that all the speculative decoding methods have the same generation quality and the acceleration ratio of our method is significantly better than other baselines in end-to-end time for all the test dataset. Using FIRP to predict the pseudo hidden states in the intermediate layer gain more benefit in the overall performance than Medusa heads to predict the token distribution directly and our draft size is almost $7$ times smaller than Medusa, this improvement is more obvious in 7B models, and we find that the acceleration ratio on Gsm8k is higher because the answer in Gsm8k is more logical and predictable with more mathematical symbols.

\begin{table*}[h]\centering
\begin{adjustbox}{}
\begin{tabular}{lcccccc}
\bottomrule
\hline
\multirow{2}{*}{\textbf{Model}} & \multicolumn{2}{c}{\multirow{2}{*}{\textbf{Decoding Algorithm}}} & \multicolumn{2}{c}{\textbf{XSum}} & \multicolumn{2}{c}{\textbf{Gsm8k}} \\
& & & Speedup & Rouge2 & Speedup & Rouge2\\ 
\hline
\multirow{4}{*}{LLaMA2-Chat-13B} & {Auto-regressive} & 0.0B & 1.000$\times$  & 0.149 & 1.000$\times$ & 0.207 \\ 
& {Self-speculative}~\cite{zhang2023draft} & 0.0B & 1.241$\times$ & 0.148 & 1.216$\times$ & 0.208\\
& {Medusa}~\cite{medusa} & 0.57B & 1.780$\times$ & 0.149 & 2.510$\times$ & 0.207 \\
& \textbf{FIRP} & 0.08B & \textbf{2.011}$\times$ & 0.149 & \textbf{2.652}$\times$ & 0.207 \\
\hline
\multirow{4}{*}{Vicuna-13B-v1.5}  & {Auto-regressive} & 0.0B & 1.000$\times$ & 0.139 & 1.000$\times$ & 0.188 \\ 
                              & {Self-speculative}~\cite{zhang2023draft} & 0.0B & 1.125$\times$ & 0.138  & 1.118$\times$  & 0.187 \\
                           & {Medusa}~\cite{medusa} & 0.57B & 1.831$\times$ & 0.138 & 2.411$\times$ & 0.186 \\
                           & {\textbf{FIRP}}  & 0.08B & \textbf{1.951}$\times$ & 0.137 & \textbf{2.623}$\times$ & 0.189 \\
\hline
\multirow{3}{*}{LLaMA2-Chat-7B}  & {Auto-regressive} & 0.0B & 1.000$\times$ & 0.125  & 1.000$\times$ & 0.180 \\ 
                           & {Medusa}~\cite{medusa} & 0.44B & 1.791$\times$ & 0.125 & 2.290$\times$ & 0.179\\
                           & {\textbf{FIRP}} & 0.05B & \textbf{2.072}$\times$ & 0.125 &\textbf{2.783}$\times$ & 0.181\\
\hline
\multirow{3}{*}{Vicuna-7B-v1.5}  & {Auto-regressive} & 0.0B & 1.000$\times$& 0.109 & 1.000$\times$ & 0.163 \\ 
                           & {Medusa}~\cite{medusa} & 0.44B & 1.681$\times$  &  0.109 & 2.162$\times$ & 0.163\\
                           & {\textbf{FIRP}}  & 0.05B & \textbf{1.982}$\times$ & 0.109 & \textbf{2.561}$\times$ & 0.163 \\
\bottomrule
\hline
\end{tabular}
\end{adjustbox}
\caption{The end-to-end time acceleration ratio and generation quality for Xsum and Gsm8k}
\label{tab:main experiment xsum gsm8k}
\vspace{-13mm}
\end{table*}

\begin{table*}[h!]\centering
\begin{adjustbox}{}
\begin{tabular}{ccc|ccc|cc|cc}
\bottomrule
\hline
\multicolumn{3}{c|}{\multirow{2}{*}{\textbf{LLaMA2-Chat-13B}}} & \multicolumn{3}{c|}{\multirow{2}{*}{\textbf{LLaMA2-Chat-7B}}} & \multicolumn{2}{c|}{\multirow{2}{*}{\textbf{Vicuna-13b-v1.5}}} & \multicolumn{2}{c}{\multirow{2}{*}{\textbf{Vicuna-7b-v1.5}}} \\
& & & & & & & &\\
\hline
LK & Me & \textbf{FIRP} & LK & Me & \textbf{FIRP} & Me & \textbf{FIRP} & Me & \textbf{FIRP} \\
1.64$\times$  & 2.34$\times$ & \textbf{2.44$\times$} & 1.51$\times$ & 2.14$\times$ & \textbf{2.29$\times$} & 2.41$\times$ & \textbf{2.45$\times$} & 2.40$\times$ & \textbf{2.49$\times$}  \\
\bottomrule
\hline
\end{tabular}
\end{adjustbox}
\caption{The end-to-end time acceleration ratio for Mt-bench.(LK, Me represent Lookahead decoding and Medusa decoding respectively)}
\label{tab:main experiment2}
\vspace{-12mm}
\end{table*}
\subsection{Analytical Study}
In this subsection, we conduct some analytical experiments to further verify the effectiveness and motivation of our method, the core idea of our method is to predict draft tokens more correctly in a single forward propagation by predicting the pseudo hidden states in intermediate layers. So we first compare our method with Medusa and early exiting to show that we have better draft tokens prediction accuracy; then we analyze how the pseudo hidden states change during the forward propagation to verify the refinement we proposed. We also verify the draft tokens prediction dependency and show how to select the prediction layers.

\subsubsection{Draft tokens prediction accuracy}
\begin{figure*}[h]
  \centering
  \includegraphics[width=1.0\textwidth, height=0.25\textwidth]{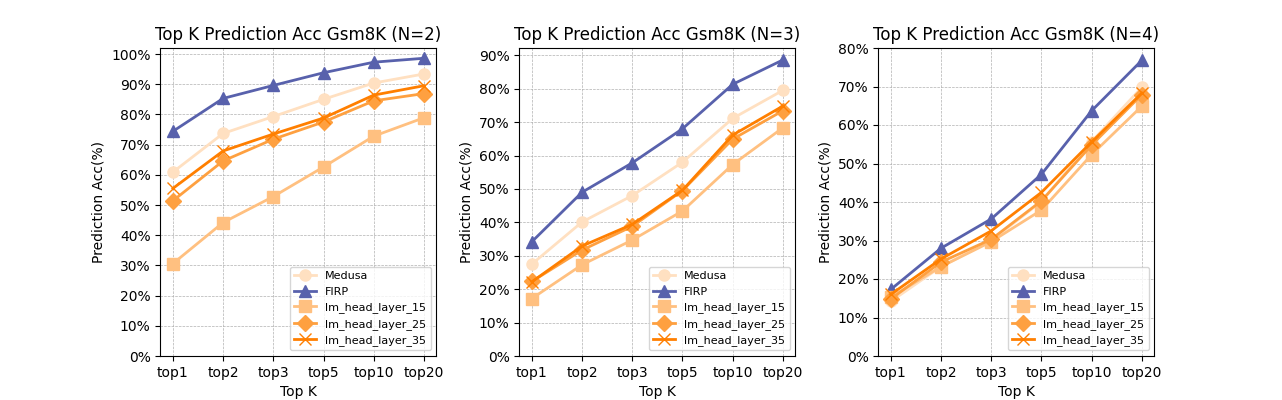}
  \caption{TopK tokens' prediction accuracy using three prediction methods on LLaMA-2-Chat-13B model including directly train different lm-heads on some intermediate layers (denoted as Early exit in the figure), Medusa method and FIRP, The $N$ in the figure is the prediction step (N=2 means we predict the first draft token). It's clear that our method achieve the best prediction accuracy}
  \label{fig:token_acc}
  \vspace{-5mm}
\end{figure*}

We first compare the draft tokens' prediction accuracy on three different methods: early exiting in the intermediate layers, using medusa heads and our method. Early exiting method trains independent lm-heads in several intermediate transformer layers to directly predict the draft tokens(for example, we can train a lm-head and use it to map $h_{n}^{j}$ to $\widetilde{X}_{n+2}$),  We use the LLaMA-2-Chat-13B as the base model for this experiment, three different methods are trained on ShareGPT dataset for one epoch and we set $K$ as 3. We random sample 100 sequences from the test split of XSum and Gsm8K dataset respectively, for each sequence, we random split 50 points at its output part, and for each split-point, we use the token sequence before as the prompt input and predict $K$ draft tokens using different methods and compare with the tokens generated greedily by the original model in the following $K$ steps.(We choose five random seeds and average the results to better eliminate random errors) Figure~\ref{fig:token_acc} shows that FIRP achieve the best prediction accuracy among them.

We also compare the average acceptance length with Medusa in Table~\ref{tab:tree structure} under different tree node number budgets, we also set $K$=3 and for each tree node number budget and search for the best tree structures for FIRP and Medusa respectively as~\cite{medusa}. The result proves that we accept more draft tokens in a single forward propagation under different tree node number budgets consistently.

\begin{table*}\centering
\begin{adjustbox}{}
\begin{tabular}{lcccccccc}
\bottomrule
\hline
\multirow{3}{*}{\textbf{Model}} & \multirow{3}{*}{\textbf{Decoding Algorithm}} & \multicolumn{3}{c}{\multirow{2}{*}{\textbf{XSum}}} & \multicolumn{3}{c}{\multirow{2}{*}{\textbf{Gsm8k}}} \\
& & & & & & & \\
& & T16& T32& T63& T16& T32& T63 \\
\hline
\multirow{2}{*}{LLaMA2-Chat-13B} & Medusa & 1.958 & 2.064 & 2.125 & 2.344 & 2.526 & 2.649\\
& \textbf{FIRP} & \textbf{2.306} &\textbf{2.449} &\textbf{2.542} &\textbf{2.698} & \textbf{2.899}& \textbf{3.054}\\
\hline
\multirow{2}{*}{Vicuna-13B-v1.5} & Medusa & 1.885 & 2.003& 2.059&2.389 & 2.569& 2.709\\
& \textbf{FIRP} &  \textbf{2.220}&  \textbf{2.359}&  \textbf{2.458}&  \textbf{2.758}&  \textbf{2.963}& \textbf{3.125} \\
\hline
\multirow{2}{*}{LLaMA2-Chat-7B} & Medusa & 1.848&1.948 & 2.000& 2.209& 2.366& 2.475\\
& \textbf{FIRP} &  \textbf{2.356}& \textbf{2.504} &  \textbf{2.622}&  \textbf{2.627}&  \textbf{2.832}&  \textbf{2.993}\\
\hline
\multirow{2}{*}{Vicuna-7B-v1.5} & Medusa & 1.835& 1.935& 1.988& 2.334& 2.513& 2.647\\
& \textbf{FIRP} & \textbf{2.377}& \textbf{2.517} &  \textbf{2.618}&  \textbf{2.762}&  \textbf{2.987}&  \textbf{3.148}\\
\bottomrule
\hline
\end{tabular}
\end{adjustbox}
\caption{Average acceptance length for Medusa and our method in one forward propagation under different tree Nodes(T16, T32, and T63 indicate that the optimal tree structure consists of 16, 32, and 63 nodes respectively)}
\label{tab:tree structure}
\vspace{-7mm}
\end{table*}

\subsubsection{Pseudo hidden states refinement}
\begin{figure}[h] 
  \centering
  \includegraphics[width=0.55\linewidth, height=0.25\textwidth]{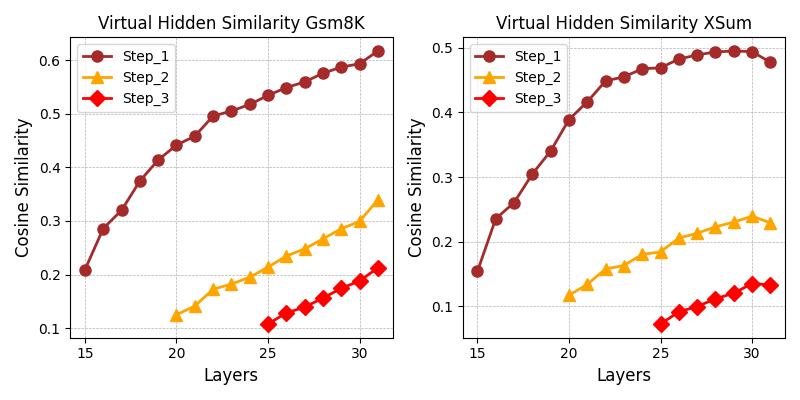}
  \caption{Hidden states similarity between the pseudo hidden states predicted and the original hidden states}
  \label{fig:hidden_similarity}
\end{figure}

 we conduct another experiment to prove that the forward propagation in transformer layers can \textbf{refine} the pseudo hidden states predicted by given more semantic information using self-attention mechanism. We compare the cosine similarity of the pseudo hidden states predicted with the \textbf{original} hidden states, and trace how the cosine similarity changes with the forward propagation (for example, we denote the prompt sequence as $X_1$, ... $X_{n}$ and we calculate the cosine similarity between pseudo hidden states $\widetilde{h}_{n+1}^{t}$ and $h_{n+1}^{t}$, $h_{n+1}^{t}$ is the hidden state of $X_{n+1}$ on the $t_{th}$ layer and $X_{n+1}$ is the greedy decoding output token given the prefix context). We random sample 100 sequences for two datasets and random split 50 times for each sequence as well. Figure~\ref{fig:hidden_similarity} shows the cosine similarity get closer along with the forward process, which proves that with the forward propagation, the hidden states can be refined by the transformer layers.

\subsubsection{How to choose prediction layers}
\begin{figure}[h]
  \centering
  \includegraphics[width=0.75\linewidth, height=0.25\linewidth]{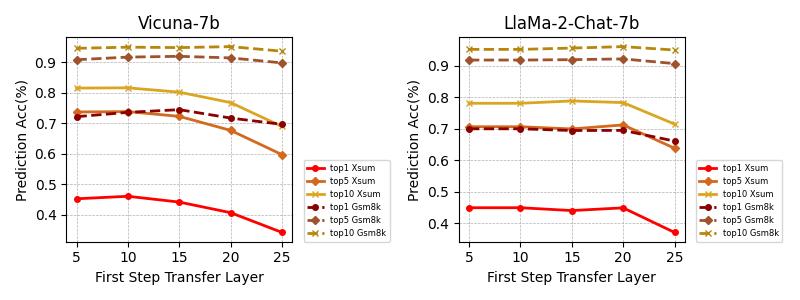}
  \caption{The first prediction step prediction accuracy on different layers for Vicuna-7b and LlaMa-2-Chat-7b.}
  \label{fig:different layers acc step1}
  \vspace{-7mm}
\end{figure}

\begin{figure}[h]
  \centering
  \includegraphics[width=0.75\linewidth, height=0.25\linewidth]{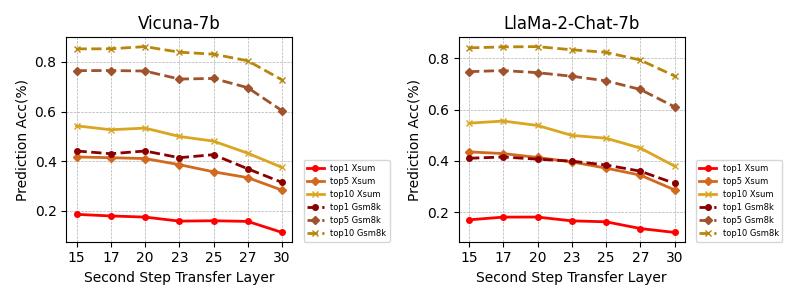}
  \caption{The second prediction step prediction accuracy on different layers for Vicuna-7b and LlaMa-2-Chat-7b with the fixed prediction step 15.}
  \label{fig:different layers acc step2}
  \vspace{-7mm}
\end{figure}

In the inference stage of our method, we need to choose which layer to predict the pseudo hidden states for each step, there's a trade-off between accuracy and efficiency: if we predict on lower layers, the pseudo hidden states will pass more subsequent layers to gain more semantic information by interacting with the context but take more computing resource; if we predict on higher layers, the pseudo hidden states will take less computing resource with less semantic information. So we train the first and second prediction step on different layers of a fixed LLM to study the impact of the prediction layer selection. We choose Vicuna-7B and LLaMA2-chat-7B as base model. For the first prediction step, we train different linear transformation on the $5_{th}$, $10_{th}$, $15_{th}$, $20_{th}$, $25_{th}$ layers respectively , and report the token prediction accuracy in figure~\ref{fig:different layers acc step1}, we found that from the lower layers to the middle layers, the prediction accuracy is basically unchanged, and from the middle layers to the high layers, the prediction accuracy drops rapidly, which proves that prediction on the middle layers is an optimal choice for both accuracy and computational efficiency, so we choose the $15_{th}$ layer to conduct the first prediction step. After fixed the first prediction layer, we also train different linear transformation on the layers higher than it as our second prediction layer. Figure~\ref{fig:different layers acc step2}
shows that the second prediction step have the same rule, starting from the $15_{th}$ layer, the accuracy remains unchanged within a range, and then rapidly decline, so we choose $20_{th}$ layer to conduct the second prediction step.

\subsection{Ablation study}
\begin{figure}[h]
  \centering
  \includegraphics[width=0.7\textwidth, height=0.3\textwidth]{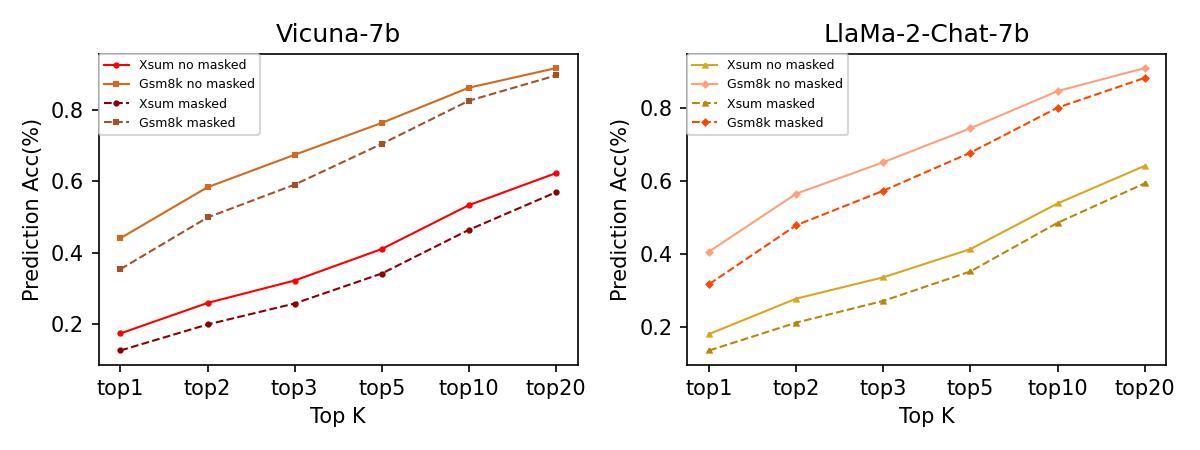}
  \caption{The second step prediction accuracy under masked and no masked setting, we predict the hidden states on $15_{th}$ and $20_{th}$ respectively}
  \label{fig:transfer_step2_need_attention.jpg}
\end{figure}

It's clear that Medusa~\cite{medusa} predict the draft tokens in parallel which means the generation between different draft tokens is independent. Our motivation is that serialized generation of draft tokens will gain better performance and we use experiment to prove it. We compare the prediction accuracy of the second prediction step under two setting: \textbf{Masked} and \textbf{No masked}, \textbf{Masked} means the second pseudo hidden state can not see the first pseudo hidden state in the forward propagation, and \textbf{No masked} is normal self-attention mechanism. Figure~\ref{fig:transfer_step2_need_attention.jpg} shows that in all the models and datasets, \textbf{Masked} performs worse which prove that the semantic information of the first draft token is important to the generation of the second draft token.
\vspace{-10mm}

\section{Conclusion}
In this paper, we introduce a novel speculative decoding approach, named FIRP, designed for accelerating inference in LLMs. FIRP is capable of predicting the pseudo hidden states of multiple future tokens in a single forward propagation. Our main experiments demonstrate that FIRP outperforms baselines in terms of prediction accuracy and also achieves substantial gains in speedup ratio at smaller draft size. Through analytical experiments, it has proved that the pseudo hidden states are progressively refined in forward propagation.


\bibliographystyle{unsrt}


\begin{thebibliography}{10}

\bibitem{touvron2023llama}
Hugo Touvron, Thibaut Lavril, Gautier Izacard, Xavier Martinet, Marie-Anne Lachaux, Timothée Lacroix, Baptiste Rozière, Naman Goyal, Eric Hambro, Faisal Azhar, Aurelien Rodriguez, Armand Joulin, Edouard Grave, and Guillaume Lample.
\newblock Llama: Open and efficient foundation language models, 2023.

\bibitem{touvron2023llama-2}
Hugo Touvron, Louis Martin, Kevin Stone, Peter Albert, Amjad Almahairi, Yasmine Babaei, Nikolay Bashlykov, Soumya Batra, Prajjwal Bhargava, Shruti Bhosale, Dan Bikel, Lukas Blecher, Cristian~Canton Ferrer, Moya Chen, Guillem Cucurull, David Esiobu, Jude Fernandes, Jeremy Fu, Wenyin Fu, Brian Fuller, Cynthia Gao, Vedanuj Goswami, Naman Goyal, Anthony Hartshorn, Saghar Hosseini, Rui Hou, Hakan Inan, Marcin Kardas, Viktor Kerkez, Madian Khabsa, Isabel Kloumann, Artem Korenev, Punit~Singh Koura, Marie-Anne Lachaux, Thibaut Lavril, Jenya Lee, Diana Liskovich, Yinghai Lu, Yuning Mao, Xavier Martinet, Todor Mihaylov, Pushkar Mishra, Igor Molybog, Yixin Nie, Andrew Poulton, Jeremy Reizenstein, Rashi Rungta, Kalyan Saladi, Alan Schelten, Ruan Silva, Eric~Michael Smith, Ranjan Subramanian, Xiaoqing~Ellen Tan, Binh Tang, Ross Taylor, Adina Williams, Jian~Xiang Kuan, Puxin Xu, Zheng Yan, Iliyan Zarov, Yuchen Zhang, Angela Fan, Melanie Kambadur, Sharan Narang, Aurelien Rodriguez, Robert Stojnic, Sergey Edunov, and Thomas
  Scialom.
\newblock Llama 2: Open foundation and fine-tuned chat models, 2023.

\bibitem{roziere2023code}
Baptiste Roziere, Jonas Gehring, Fabian Gloeckle, Sten Sootla, Itai Gat, Xiaoqing~Ellen Tan, Yossi Adi, Jingyu Liu, Tal Remez, J{\'e}r{\'e}my Rapin, et~al.
\newblock Code llama: Open foundation models for code.
\newblock {\em arXiv preprint arXiv:2308.12950}, 2023.

\bibitem{shazeer2019fast}
Noam Shazeer.
\newblock Fast transformer decoding: One write-head is all you need.
\newblock {\em arXiv preprint arXiv:1911.02150}, 2019.

\bibitem{leviathan2023fast}
Yaniv Leviathan, Matan Kalman, and Yossi Matias.
\newblock Fast inference from transformers via speculative decoding.
\newblock In {\em International Conference on Machine Learning}, pages 19274--19286. PMLR, 2023.

\bibitem{chen2023accelerating}
Charlie Chen, Sebastian Borgeaud, Geoffrey Irving, Jean-Baptiste Lespiau, Laurent Sifre, and John Jumper.
\newblock Accelerating large language model decoding with speculative sampling.
\newblock {\em arXiv preprint arXiv:2302.01318}, 2023.

\bibitem{zhang2023draft}
Jun Zhang, Jue Wang, Huan Li, Lidan Shou, Ke~Chen, Gang Chen, and Sharad Mehrotra.
\newblock Draft \& verify: Lossless large language model acceleration via self-speculative decoding.
\newblock {\em arXiv preprint arXiv:2309.08168}, 2023.

\bibitem{medusa}
Tianle Cai, Yuhong Li, Zhengyang Geng, Hongwu Peng, and Tri Dao.
\newblock Medusa: Simple framework for accelerating llm generation with multiple decoding heads.
\newblock \url{https://github.com/FasterDecoding/Medusa}, 2023.

\bibitem{miao2023specinfer}
Xupeng Miao, Gabriele Oliaro, Zhihao Zhang, Xinhao Cheng, Zeyu Wang, Rae Ying~Yee Wong, Zhuoming Chen, Daiyaan Arfeen, Reyna Abhyankar, and Zhihao Jia.
\newblock Specinfer: Accelerating generative llm serving with speculative inference and token tree verification.
\newblock {\em arXiv preprint arXiv:2305.09781}, 2023.

\bibitem{spector2023accelerating}
Benjamin Spector and Chris Re.
\newblock Accelerating llm inference with staged speculative decoding.
\newblock {\em arXiv preprint arXiv:2308.04623}, 2023.

\bibitem{fu2024break}
Yichao Fu, Peter Bailis, Ion Stoica, and Hao Zhang.
\newblock Break the sequential dependency of llm inference using lookahead decoding, 2024.

\bibitem{Liu2024SpeculativeDV}
Jiahao Liu, Qifan Wang, Jingang Wang, and Xunliang Cai.
\newblock Speculative decoding via early-exiting for faster llm inference with thompson sampling control mechanism.
\newblock {\em ArXiv}, abs/2406.03853, 2024.

\bibitem{Gong2024GraphStructuredSD}
Zhuocheng Gong, Jiahao Liu, Ziyue Wang, Pengfei Wu, Jingang Wang, Xunliang Cai, Dongyan Zhao, and Rui Yan.
\newblock Graph-structured speculative decoding.
\newblock 2024.

\bibitem{vicuna2023}
Wei-Lin Chiang, Zhuohan Li, Zi~Lin, Ying Sheng, Zhanghao Wu, Hao Zhang, Lianmin Zheng, Siyuan Zhuang, Yonghao Zhuang, Joseph~E. Gonzalez, Ion Stoica, and Eric~P. Xing.
\newblock Vicuna: An open-source chatbot impressing gpt-4 with 90\%* chatgpt quality, March 2023.

\bibitem{wolf2020transformers}
Thomas Wolf, Lysandre Debut, Victor Sanh, Julien Chaumond, Clement Delangue, Anthony Moi, Pierric Cistac, Tim Rault, R{\'e}mi Louf, Morgan Funtowicz, et~al.
\newblock Transformers: State-of-the-art natural language processing.
\newblock In {\em Proceedings of the 2020 conference on empirical methods in natural language processing: system demonstrations}, pages 38--45, 2020.

\bibitem{lhoest2021datasets}
Quentin Lhoest, Albert~Villanova del Moral, Yacine Jernite, Abhishek Thakur, Patrick von Platen, Suraj Patil, Julien Chaumond, Mariama Drame, Julien Plu, Lewis Tunstall, et~al.
\newblock Datasets: A community library for natural language processing.
\newblock {\em arXiv preprint arXiv:2109.02846}, 2021.

\bibitem{narayan2018don}
Shashi Narayan, Shay~B Cohen, and Mirella Lapata.
\newblock Don't give me the details, just the summary! topic-aware convolutional neural networks for extreme summarization.
\newblock {\em arXiv preprint arXiv:1808.08745}, 2018.

\bibitem{cobbe2021training}
Karl Cobbe, Vineet Kosaraju, Mohammad Bavarian, Mark Chen, Heewoo Jun, Lukasz Kaiser, Matthias Plappert, Jerry Tworek, Jacob Hilton, Reiichiro Nakano, et~al.
\newblock Training verifiers to solve math word problems.
\newblock {\em arXiv preprint arXiv:2110.14168}, 2021.

\end{thebibliography}

\end{document}